\documentclass[10pt,twocolumn,letterpaper]{article}
\usepackage[pagenumbers]{cvpr}
\usepackage{multirow}
\usepackage{array}
\usepackage{placeins}
\usepackage{float}
\newlength{\openlooptablewidth}

\definecolor{linkblue}{rgb}{0.21,0.49,0.74}
\usepackage[breaklinks,colorlinks,allcolors=linkblue]{hyperref}

\title{GraF-VTA: Neuro-Symbolic Trajectory Generation from Vision and Temporal Logic via Graph-Based Fusion}

\author{
Zezhi Liu\textsuperscript{1,2}\textsuperscript{*},
Zhiwei Zheng\textsuperscript{1}\textsuperscript{*},
Hanqian Luo\textsuperscript{1,3},
Deyun Qin\textsuperscript{1},\\
Shizhen Wu\textsuperscript{1},
Yongchun Fang\textsuperscript{1}\textsuperscript{\textdagger}\\[0.6em]
{\small \textsuperscript{1} Nankai University}\\
{\small \textsuperscript{2} Zhongguancun Academy}\\
{\small \textsuperscript{3} The Hong Kong Polytechnic University}\\
{\footnotesize \textsuperscript{*} Equal contribution. \quad
\textsuperscript{\textdagger} Corresponding author.}
}

\begin{document}
\maketitle
\begin{abstract}
Robotic planning from visual observations and signal temporal logic (STL) requires grounding task requirements in scene evidence while adapting plans to execution feedback. We present GraF-VTA, a graph-based framework that extends Vision-TL-Action with structured task representations, causal observation history, and closed-loop trajectory generation. Bidirectional cross-attention fuses STL syntax tokens with visual features, while an optional frozen automaton embedding supplies complementary untimed task structure. A temporal pyramid preserves detailed current observations and compresses older frames to provide recent execution context. Conditioned on these features and robot state, a flow-matching model generates plans whose action prefixes are executed before replanning from refreshed observations. Across five simulation environments, the single-graph configuration achieves higher open-loop Success@$K$ than TeLoGraF and Vision-TL-Action on four environments at the reported budgets $K\geq16$. On Panda, matched-update ablations increase Success@256 from 53.91\% without observation history to 62.50\% with pyramid memory. In closed-loop Dubins evaluation, it reaches 64.06\% Success@1024, compared with 12.50\% for adapted S-MSP and 27.34\% for adapted Logic-VLA. These results support combining structured visual conditioning with observation history and replanning to improve coverage of STL-satisfying trajectories.
\end{abstract}

\section{Introduction}
\label{sec:intro}
Robotic tasks often combine goal achievement with deadlines, ordering, and obstacle avoidance. Signal temporal logic (STL) expresses these requirements precisely through spatial predicates and bounded temporal operators~\citep{maler2004monitoring,belta2019formal,lindemann2019feedback}. Trajectory generation and execution from visual observations, STL specifications, and robot state are therefore important for fulfilling these requirements.

Visual input provides a practical interface between STL planning and robot sensing, with grounding mechanisms relating symbolic references to image regions~\citep{Kamath_2021_ICCV,Li_2022_CVPR,liu2024grounding}. 
Pretrained vision-language models have also been used to propose visual predicates and evaluate them from images, enabling the learning of symbolic world models for long-horizon planning~\cite{athalye2026p2p}.
For STL-conditioned trajectory generation, TeLoGraF~\citep{meng2025telograf} conditions trajectory generation on syntax graphs containing exact object geometry, which must be supplied externally. End-to-end visuomotor learning~\citep{levine2016endtoend,brohan2023rt,zitkovich2023rt2,li2024roboflamingo} demonstrates that perception can instead be trained jointly with action prediction. Vision-TL-Action (VTA)~\citep{liu2026visiontlaction} brings visual grounding to STL-conditioned generation by connecting logic nodes to image tokens. This supports a learned observation-to-trajectory mapping without exact object geometry in the STL specifications.

Executing trajectories planned by generative models is challenging because model mismatch, disturbances, and tracking errors can separate intended and realized motion. Receding-horizon STL control~\citep{sadraddini2015robust,raman2015reactive,meng2023signal} and Diffusion Policy~\citep{chi2025diffusion} motivate repeated planning from execution feedback. We therefore execute short action prefixes and refresh visual observations and robot state before generating subsequent actions, allowing plans to respond to observed deviations.

Refreshed observations alone may not reveal the robot's recent interactions with the environment or the current task stage. MemoryVLA~\citep{shi2026memoryvla} demonstrates the value of historical perceptual and semantic information for temporally dependent manipulation. This motivates conditioning STL planning on causal observation history. A temporal pyramid preserves detailed current observations while compressing older frames, providing recent execution context within a finite memory window.

A complementary challenge is exposing task semantics to the generator. An STL syntax graph preserves composition, time bounds, and object references. Yet, as Xie et al.~\citep{xie2021embedding} observe for finite-trace LTL, equivalent formulas can have different syntax trees, complicating semantic learning. A deterministic finite automaton (DFA) makes the accepted event sequences explicit through transitions and accepting states. This motivates combining timed, visually groundable syntax with a semantic view.

We introduce GraF-VTA, a unified visual temporal-logic planning framework supporting STL syntax graph and DFA graph conditioning (Fig.~\ref{fig:dvta_overview}). It also combines bidirectional logic--vision fusion, causal observation memory, and action-prefix replanning within a conditional flow-matching generator. The optional dual-graph configuration adds a frozen RADE embedding~\citep{yalcinkaya2024compositional} after pooling and base-condition encoding. Our main empirical analysis concerns the single-graph configuration; dual-graph conditioning is a complementary representation extension.

Our contributions are threefold:
(1) We present a unified framework for end-to-end trajectory generation from visual observations, STL specifications, and robot state. Building on TeLoGraF's graph-conditioned flow matching and VTA's visual grounding, bidirectional logic--vision fusion connects object references to scene evidence while retaining temporal constraints. Conditioning uses the STL syntax tree alone or adds a DFA embedding for complementary task-transition structure.
(2) To address execution errors, we interleave action-prefix execution with replanning from updated visual observations and robot state. A causal temporal pyramid retains detailed current observations while compressing older frames, providing recent execution context for multi-stage tasks. This feedback-driven design responds to observed deviations, aiming to improve robustness to execution errors.
(3) We evaluate open-loop trajectory generation and dynamics-executed rollouts across five simulation environments, complemented by comparisons with related planners and observation-memory ablations. These experiments demonstrate the effectiveness of the proposed method. 

\section{Related Work}
\paragraph{Temporal-logic-conditioned robot planning.}
Temporal-logic planners incorporate specifications through constrained optimization~\citep{dawson2022robust}, sampling guidance~\citep{vasile2017sampling}, or learned conditioning~\citep{guo2024specification}. Differentiable STL robustness supports optimization~\citep{leung2023backpropagation,kapoor2025stlcg}; generative planners use STL-constrained diffusion~\citep{meng2024diverse,zhong2023guided}, hierarchical options~\citep{feng2025diffusion}, or trajectory composition~\citep{liu2026zeroshotstl}. Robust STL model predictive control~\citep{sadraddini2015robust} uses mixed-integer optimization under linear dynamics and bounded disturbances, whereas LTLDoG~\citep{feng2024ltldog} guides diffusion sampling with finite-trace LTL satisfaction at inference time. TeLoGraF~\citep{meng2025telograf} instead learns a flow-matching generator conditioned on an STL syntax graph with exact object geometry. VTA~\citep{liu2026visiontlaction} replaces that geometric input with node-level graph--vision cross-attention and training-time predicate-to-region supervision. Among other visual STL planners, S-MSP~\citep{ye2026smsp} combines images, STL structure, and pose--region context in a mixture-of-experts waypoint decoder, with RRT-based local trajectory repair. Logic-VLA~\citep{wang2026logicvla} pretrains an STL encoder using robustness supervision and adapts a pretrained VLA through supervised fine-tuning and trajectory-level preference optimization. Our framework retains VTA's visual--symbolic interface and flow-matching formulation, extending its conditioning with causal observation history and optional automaton features with closed-loop replanning.

\vspace{-1.4em}
\paragraph{Learning representations of logic and automata.}
Learned task representations differ in the structures they encode and how they influence downstream behavior. LTL2Action~\citep{vaezipoor2021ltl2action} combines structured LTL instructions with environment-agnostic pretraining for multi-task reinforcement learning. T-LEAF~\citep{xie2021embedding} embeds DFAs derived from finite-trace LTL using a hierarchical graph encoder trained with satisfying and violating traces; distances in this space define a logic loss for sequential prediction and imitation learning. RADE~\citep{yalcinkaya2024compositional} pretrains graph embeddings on reach--avoid-derived automata for reinforcement learning conditioned on compositional DFAs. We reuse frozen RADE features to condition trajectory generation alongside the visually grounded STL representation. Its embedding represents the original component initial states, fixed throughout execution rather than tracking observed task progress. The STL representation retains time bounds and object references, and the original timed specification remains the evaluation criterion.

\vspace{-1.4em}
\paragraph{Visual generative policies and observation memory.}
Generative planners represent multiple possible behaviors through distributions over trajectories or action sequences~\citep{black2024pi0,ghosh2024octo,bjorck2025groot}. Diffuser~\citep{janner2022planning} plans by denoising trajectories, while Diffusion Policy~\citep{chi2025diffusion} conditions action-sequence denoising on visual observations and executes predictions in a receding horizon; FlowPolicy~\citep{zhang2025flowpolicy} uses consistency flow matching. VLA research also explores multimodal prompts~\citep{jiang2023vima}, cross-embodiment learning~\citep{openx2024,kim2025openvla}, spatial representations~\citep{zhen2024threedvla,qu2025spatialvla}, open-world generalization~\citep{pi2025pi05}, and simplified architectures~\citep{ye2026starvlaalpha}. Recent VLA models retain history in different ways, with MemoryVLA~\citep{shi2026memoryvla} retrieving and consolidating perceptual and semantic features, HAMLET~\citep{koo2026hamlet} aggregating compressed moment tokens across time, and MEM~\citep{torne2026mem} combining short-term video memory with long-term textual memory. Our planner uses a finite window of visual observations with an explicit allocation of spatial tokens such that current frames retain fine detail, while older frames receive coarser representations. Both current and historical tokens participate in bidirectional fusion with STL nodes before conditioning flow matching. This observation history does not reuse previously generated plans, and removing it still leaves closed-loop replanning.

\section{Method}
\begin{figure*}[t]
    \centering
    \includegraphics[width=\textwidth]{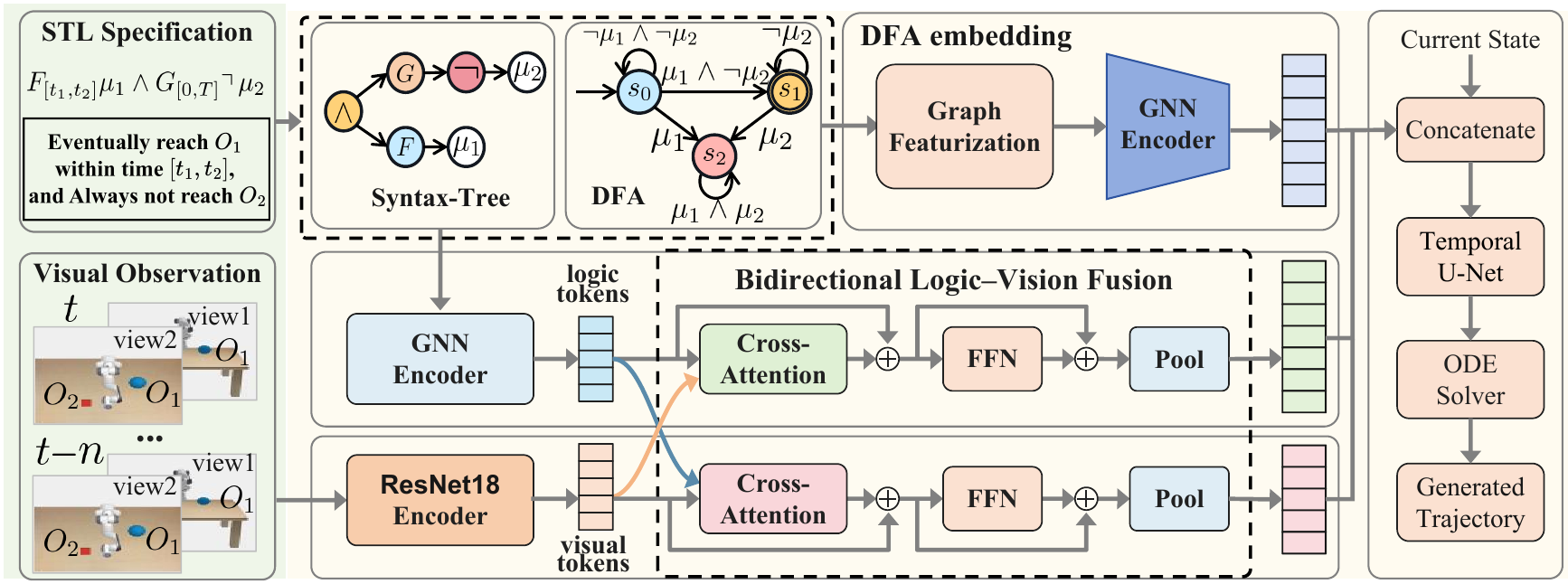}
    \caption{\textbf{System diagram for GraF-VTA}. The STL specification is represented as a syntax tree and a deterministic finite automaton (DFA), both encoded by GNNs. Logic tokens from the syntax tree and visual tokens from multi-view observations are fused through bidirectional cross-attention. The fused features, DFA embedding, and current state condition a Temporal U-Net to generate the planned trajectory via an ODE solver.}
    \label{fig:dvta_overview}
\end{figure*}

\subsection{Problem Formulation}
\label{sec:problem}

Let 
$\boldsymbol{\tau}=\left(
\boldsymbol{q}_0 , \boldsymbol{a}_0, \boldsymbol{q}_1, \boldsymbol{a}_1, \ldots , \boldsymbol{q}_{T-1} , \boldsymbol{a}_{T-1}\right)$ 
denote a trajectory, where
$\boldsymbol{q}_t\in\mathbb{R}^n$ and
$\boldsymbol{a}_t\in\mathbb{R}^m$ denote the system state and action at time $t$,
respectively, and $T$ is the planning horizon. 
For a trajectory $\boldsymbol\tau$, we denote its temporally aligned multi-view observation sequence by $\mathcal O=\left(\mathcal{O}_t\right)_{t=0}^{T-1}$, where $\mathcal{O}_t=\{O_t^h\}_{h=1}^{H}$ contains the $H$ camera views observed at time $t$.
STL specifies tasks through state predicates and temporal operators~\citep{lindemann2021funnel}.
An atomic predicate $\mu$ holds when $h_\mu(\boldsymbol{q}_t)>0$.
Formulas are defined recursively by
\begin{equation}
    \Phi ::= \top \mid \mu \mid \neg\Phi
    \mid \Phi_1\wedge\Phi_2
    \mid \Phi_1\,U_{[a,b]}\,\Phi_2,
    \label{eq:stl_syntax}
\end{equation}
where $\top$, $\neg$, and $\wedge$ denote true, negation, and conjunction,
and $U_{[a,b]}$ denotes until over the interval $[a,b]$, with $0\leq a\leq b$.
Eventually and always are derived as
$F_{[a,b]}\Phi=\top\,U_{[a,b]}\,\Phi$ and
$G_{[a,b]}\Phi=\neg F_{[a,b]}\neg\Phi$.
We write $(\boldsymbol{\tau},t)\models\Phi$ when the trajectory satisfies $\Phi$ at time $t$.

We consider a dataset of demonstrations
$\mathcal{D}=\left\{\left(\Phi_i,
\{\boldsymbol{\tau}_{i,j}, \mathcal{O}_{i,j}\}_{j=1}^{R}\right)\right\}_{i=1}^{W}$,
where $\Phi_i$ is the STL formula defined in \eqref{eq:stl_syntax}, $\mathcal{O}_{i,j}$ is the multi-view observation sequence
temporally aligned with the $j$-th demonstrated trajectory
$\boldsymbol{\tau}_{i,j}$ of task $i$,
and $\{\boldsymbol{\tau}_{i,j}\}_{j=1}^{R}$ is a set of demonstrated trajectories satisfying the specification, i.e., \((\boldsymbol{\tau}_{i,j},0)\models\Phi_i\) for all \(j\in\{1,\ldots,R\}\). 
 Given an STL specification $\Phi$, let $\mathcal{G}_{\Phi}$ denote the collection of coordinate-free graphs derived from $\Phi$.
At planning time $t$, the observation buffer is defined as $\mathcal{B}_t=(\mathcal{O}_s)_{s\in\mathcal{T}_t}$, $\mathcal{T}_t\subseteq\{0,\ldots,t\}$, $t\in\mathcal{T}_t$,
where $\mathcal{T}_t$ indexes the observations retained from
$\mathcal{O}$, including the current observation.
Without observation history, $\mathcal{T}_t=\{t\}$
and $\mathcal{B}_t=(\mathcal{O}_t)$.
We aim to learn a conditional generative planner
$\boldsymbol{y}_t \sim
    p_{\theta}\!\left(
        \cdot \mid
        \boldsymbol{q}_t,\mathcal{B}_t,\mathcal{G}_{\Phi}
    \right),$
where $\boldsymbol{y}_t$ denotes a future plan.
The planner is queried once at $t=0$ for open-loop generation,
or repeatedly with updated observations and robot state
for closed-loop execution.
The objective is to produce a full trajectory
$\boldsymbol{\tau}$ satisfying the original timed specification,
$(\boldsymbol{\tau},0)\models\Phi$, where $\boldsymbol{\tau}$
is the generated trajectory in open loop or the realized
trajectory in closed loop. STL satisfaction is evaluated on the complete trajectory using the original episode time origin. Observation age identifies the relative age of retained frames and does not itself encode elapsed task time. Details of task-time conditioning and the treatment of STL temporal bounds during replanning are provided in Appendix.

\subsection{Overview}
Figure~\ref{fig:dvta_overview} summarizes the framework. The STL graph and causal visual history are encoded into tokens, fused bidirectionally, and pooled. The pooled features and encoded robot state condition trajectory generation. Syntax-only conditioning forms the base configuration; an optional frozen RADE embedding adds an automaton view after base-condition encoding. A temporal U-Net learns a conditional flow from noise to plans. During closed-loop execution, the planner executes an action prefix and refreshes its observations and state before generating the next plan.

\subsection{STL Graph Encoding}
\label{sec:stl_graph_encoding}
To capture the structure of $\Phi$, we treat it as a syntax tree with one node per operator or atomic predicate (AP) occurrence.
The resulting graph is $G_\Phi^{\mathrm{syn}}=(\mathcal V_\Phi,\mathcal E_\Phi)$, where $\mathcal V_\Phi$ contains the nodes and $\mathcal E_\Phi$ records parent--child connections.
Each node has a feature vector $\boldsymbol h_v=\operatorname{LN}(\boldsymbol e_v^{\mathrm{cat}}+\boldsymbol e_v^{\mathrm{time}})\in\mathbb R^d$, where $d=256$ and $\operatorname{LN}$ denotes layer normalization.
Here, $\boldsymbol e_v^{\mathrm{cat}}$ sums learned embeddings of valid categorical attributes: node type, predicate label, object slots and types, argument positions, per-object occurrence index, depth, and child order.
The temporal feature $\boldsymbol e_v^{\mathrm{time}}$ is a two-layer MLP projection of normalized bounds $(a_v/T,b_v/T)$, or zero when no interval is available.
Object slots are scene-local symbolic references without coordinates or radii; child order distinguishes the operands of $U_{[a,b]}$.

A multi-layer graph neural network (GNN) takes $G_\Phi^{\mathrm{syn}}$ and the node-feature matrix $\boldsymbol H_\Phi\in\mathbb R^{|\mathcal V_\Phi|\times d}$, formed by stacking $\boldsymbol h_v^\top$, as input.
Starting from $\boldsymbol z_v^{(0)}=\boldsymbol h_v$, each layer updates node representations over bidirectional syntax edges and self-loops:
\begin{equation}
 \begin{aligned}
 \boldsymbol m_v^{(\ell)}
 &=\frac{\sum_{u\in\mathcal N(v)\cup\{v\}}\boldsymbol z_u^{(\ell)}}{|\mathcal N(v)|+1},\\
 \boldsymbol z_v^{(\ell+1)}
 &=\operatorname{LN}\!\left(\boldsymbol z_v^{(\ell)}+
 f_\ell([\boldsymbol z_v^{(\ell)};\boldsymbol m_v^{(\ell)}])\right),
 \end{aligned}
 \label{eq:stl_message_passing}
\end{equation}
where $\mathcal N(v)$ contains the parent and children of $v$, $f_\ell$ is a two-layer MLP with GELU and dropout, and $[\cdot;\cdot]$ denotes feature concatenation.
After $L$ layers, the GNN outputs $\boldsymbol Z_\Phi=\mathcal F_{\mathrm{syn}}(G_\Phi^{\mathrm{syn}},\boldsymbol H_\Phi)\in\mathbb R^{|\mathcal V_\Phi|\times d}$, whose rows are $(\boldsymbol z_v^{(L)})^\top$.
These node embeddings serve as STL tokens for subsequent logic--vision fusion.

As an optional extension, we compile an untimed abstraction of $\Phi$ into a conjunctive composition of deterministic finite automata (cDFA), sharing propositions for identical predicate events.
Pretrained RADE represents automaton states and labeled transitions as graph nodes and connects component initial states to a conjunction root.
GATv2 message passing and root readout produce $\boldsymbol r_\Phi\in\mathbb R^{32}$.
We freeze RADE and keep its input initial states fixed throughout execution.
This static embedding supplies untimed transition structure for global conditioning; temporal bounds remain in the STL branch.

\subsection{Causal Temporal-Pyramid Visual Encoding}
We encode $L_{\rm mem}$ observation slots from $\mathcal B_t$ at fixed intervals $\Delta$, with age $\ell$ corresponding to $t-\ell\Delta$ and $\ell=0$ denoting the current observation. A shared ResNet18 encoder~\citep{he2016deep} extracts $P$ tokens per view. Before pooling, fixed sinusoidal tags encode spatial position, view identity, and observation age.

The temporal pyramid allocates a total token budget across the $H$ views:
\begin{equation}
 B_\ell=H\max\!\left(1,\left\lfloor\frac{P}{2^\ell}\right\rfloor\right),
 \qquad \ell=0,\ldots,L_{\rm mem}-1.
 \label{eq:temporal_pyramid_budget}
\end{equation}
Current tokens remain unpooled, while older observations undergo spatial average pooling independently within each view. Concatenating these tokens yields $V_t$, preserving current detail and coarser historical context. Each slot's validity is expanded to a token mask $M_t$, excluding unavailable history from subsequent attention and pooling.

\subsection{Bidirectional Logic--Vision Fusion}
Bidirectional cross-attention grounds STL tokens in $V_t$ and conditions visual tokens on the task. In each layer, STL queries attend to visual keys and values, and visual queries attend to STL keys and values, both using the layer inputs. Residual attention and feed-forward updates with layer normalization refine both streams.

Validity masks exclude padded logic tokens and unavailable observations from attention and pooling. Masked mean pooling yields logic and visual vectors $\bar{\boldsymbol x}_t$ and $\bar{\boldsymbol v}_t$. In the single-graph configuration, the generator condition is
\begin{equation}
\boldsymbol c_t=f_c([\bar{\boldsymbol x}_t;\bar{\boldsymbol v}_t;f_q(\boldsymbol q_t)]),
\label{eq:dvta_condition}
\end{equation}
where $f_q$ encodes robot state, $f_c$ projects the concatenated features, and semicolons denote concatenation. The dual-graph extension incorporates $\boldsymbol r_\Phi$ at the global conditioning stage, outside token-level fusion.

\subsection{Conditional flow for trajectory generation}
\label{sec:conditional_flow}
Given the global condition constructed above, a temporal U-Net $v_\theta$ learns to generate the future plan $\boldsymbol y_t$ by conditional flow matching~\citep{lipman2023flow}.
We denote this condition by $\tilde{\boldsymbol c}_t$, using $\boldsymbol c_t$ from Eq.~\eqref{eq:dvta_condition} for the single-graph configuration and the DFA-augmented representation for the dual-graph configuration.
For each training sample from $\mathcal D$, let $\boldsymbol X_1$ be the normalized demonstration target aligned with $(\boldsymbol q_t,\mathcal B_t,\mathcal G_\Phi)$.
The target contains state--action pairs for trajectory prediction or actions for the action-only interface.
We independently sample Gaussian noise $\boldsymbol X_0\sim\mathcal N(0,I)$ of the same shape and flow time $s\sim\mathcal U(0,1)$, distinct from execution time $t$.
The interpolation $\boldsymbol X_s=(1-s)\boldsymbol X_0+s\boldsymbol X_1$ has target velocity $\boldsymbol X_1-\boldsymbol X_0$, giving the loss
\begin{equation}
\mathcal L_{\rm FM}=\mathbb E\!\left[
\left\|v_\theta(\boldsymbol X_s,s,\tilde{\boldsymbol c}_t)
-(\boldsymbol X_1-\boldsymbol X_0)\right\|_w^2
\right],
\label{eq:dvta_flow}
\end{equation}
where the expectation is over training samples, noise, and flow time, and $\|\cdot\|_w^2$ averages squared error over valid target entries.
This objective updates the velocity field and trainable conditioning modules.

The learned velocity field defines a flow $\psi_s$ through $d\psi_s(\boldsymbol X)/ds=v_\theta(\psi_s(\boldsymbol X),s,\tilde{\boldsymbol c}_t)$, with initial condition $\psi_0(\boldsymbol X)=\boldsymbol X$ and $\tilde{\boldsymbol c}_t$ fixed for each planning call.
At inference, we draw fresh noise $\boldsymbol X'\sim\mathcal N(0,I)$ and obtain the normalized plan $\psi_1(\boldsymbol X')$ by integrating from $s=0$ to $s=1$.
With $N_s$ uniform steps and $s_i=i/N_s$, Euler's method gives $\psi_{s_{i+1}}(\boldsymbol X')=\psi_{s_i}(\boldsymbol X')+v_\theta(\psi_{s_i}(\boldsymbol X'),s_i,\tilde{\boldsymbol c}_t)/N_s$ for $i=0,\ldots,N_s-1$.
The endpoint is decoded using the corresponding output normalization to obtain $\boldsymbol y_t$.

\section{Experiments}
\label{sec:experiments}

\paragraph{Implementation details.}
We evaluate our method in the five robotic simulation environments used in TeLoGraF~\citep{meng2025telograf}: Linear and Dubins, based on ZoneEnv~\citep{jackermeier2025deepltl} with single-integrator and car dynamics, respectively; PointMaze and AntMaze~\citep{fu2020d4rl}; and the Franka Panda robotic arm~\citep{gaz2019dynamic}. We apply fixes to address issues in the original AntMaze environment, and all reported AntMaze results are obtained with the same fixes applied to every method. The trajectory length is 512 for PointMaze and AntMaze and 64 for the other environments.
GraF-VTA conditions on two visual inputs and the robot state. On Panda, the visual inputs are the bird’s-eye and custom camera views, while others serves only same visual inputs. During execution, all of the visual inputs are updated in the observations.
GraF-VTA builds on the U-Net architecture ~\citep{janner2022planning}, and incorporates a visual encoder~\citep{he2016deep}, a temporal memory module for observation-history conditioning~\citep{chi2025diffusion}, bidirectional logic--vision cross-attention~\citep{liu2026visiontlaction}, and an optional DFA embedding branch.
In the single-graph configuration, the pooled STL and visual embeddings are concatenated with the robot-state encoding produced by the state encoder to form the additional conditioning inputs to the U-Net.
In the dual-graph configuration, these inputs are formed by concatenating the pooled, visually grounded STL embedding, the robot-state encoding, and the DFA embedding.
For the DFA branch, we use a pretrained RADE encoder~\citep{yalcinkaya2024compositional}, whose pretraining dataset is randomly generated.
The learning pipeline therefore uses the pretrained DFA encoder as a fixed feature extractor while learning the trajectory generator with the flow-matching objective~\citep{lipman2023flow}.
The learning pipeline is implemented in PyTorch and PyTorch Geometric. Each training job runs on NVIDIA A100 Tensor Core GPU. Evaluation uses 128 STL specification tasks per environment.

\vspace{-1.4em}
\paragraph{Baselines.}
We compare with four planners for STL specification.
For open-loop evaluation, \textbf{TeLoGraF}~\citep{meng2025telograf} uses STL graphs with exact geometry for flow-based planning, while  \textbf{Vision-TL-Action} (VTA)~\citep{liu2026visiontlaction} grounds STL graphs in multi-view images.
For closed-loop evaluation on Dubins, we adapt \textbf{S-MSP}~\citep{ye2026smsp}, a mixture-of-experts planner with candidate-wise MC dropout, and \textbf{Logic-VLA}~\citep{wang2026logicvla}, an STL-conditioned $\pi_{0.5}$ policy~\citep{pi2025pi05} using trajectory-level preference optimization.
In the ablation study, we compare GraF-VTA in closed loop with \textbf{no history}, \textbf{standard two-frame memory} with equal per-frame token budgets, and \textbf{pyramid two-frame memory} with compressed history. We also include a \textbf{dual-graph variant} of GraF-VTA that additionally uses a DFA graph. The comparisons concern the Dubins adaptations are shown in the Appendix part.

\vspace{-1.4em}
\paragraph{Metrics.}
For each task $i$ with STL specification $\Phi_i$,
we sample $K$ candidate trajectories
$\{\tau_{i,j}\}_{j=1}^{K}$ and evaluate STL robustness
$\rho^{\Phi_i}(\tau_{i,j},0)$ offline.
We report the best-of-$K$ success rate, expressed as a
percentage:
\begin{equation*}
\operatorname{Success}@K =
\frac{100}{N}\sum_{i=1}^{N}
\mathbf{1}\!\left\{
\max_{1\leq j\leq K}
\rho^{\Phi_i}(\tau_{i,j},0)>0
\right\}.
\label{eq:open_loop_success}
\end{equation*}
Here, $N$ is the number of tasks, $\tau_{i,j}$ is the
$j$th candidate for task $i$,
$\rho^{\Phi_i}(\tau_{i,j},0)$ denotes its STL robustness
evaluated at the initial time, and $\mathbf{1}\{\cdot\}$ is
the indicator function.
A task is counted as successful if at least one of its
$K$ candidates has strictly positive STL robustness.
This criterion is equivalent to evaluating the candidate
with the highest STL robustness.

\begin{figure}[H]
    \centering
    \includegraphics[width=\linewidth]{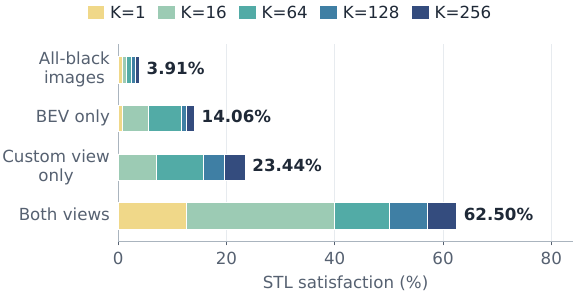}
    \caption{\textbf{Dual-view ablation of GraF-VTA with pyramid memory on Panda.}We compare both views, bird’s-eye view (BEV) only, custom view only, and all-black image inputs. Stacked segments show incremental gains in STL satisfaction as the candidate budget \(K\) increases, with endpoint labels reporting Success@256.}
    \label{fig:figure2_panda_visual_stacked_single_column}
\end{figure}
\begin{table*}[!t]
\centering
\caption{\textbf{Observation-memory and graph-representation ablations on Panda.} Success@256 and category-wise satisfaction follow the best-of-256 criterion, with one generation seed per configuration. VTA uses open-loop planning, which serves as a system-level reference; the observation-memory comparison is between the GraF-VTA configurations; GraF-VTA variants use closed-loop replanning. The dual-graph variant additionally incorporates a DFA embedding from a frozen encoder. Standard and pyramid memory both use two observation frames. I--IV denote single-goal, multi-goal, sequential, and partial-order tasks. Bold indicates the best result in each column, including ties.}
\label{tab:panda_memory_comparison}
\small
\setlength{\tabcolsep}{4pt}
\renewcommand{\arraystretch}{1.4}
\setlength{\extrarowheight}{0pt}
\setlength{\aboverulesep}{1.5pt}
\setlength{\belowrulesep}{1.5pt}

\begin{tabular*}{\textwidth}
{@{\extracolsep{\fill}}lc*{4}{>{\centering\arraybackslash}p{32pt}}c@{}}
\toprule
\multirow{2}{*}{Method}
& \multirow{2}{*}{\shortstack{Success@256\\(\%) $\uparrow$}}
& \multicolumn{4}{c}{STL satisfaction rate (\%) $\uparrow$}
& \multirow{2}{*}{\shortstack{Runtime\\(s) $\downarrow$}} \\
\cmidrule(lr){3-6}
& & I & II & III & IV & \\
\midrule
\shortstack[l]{VTA} & 50.00 & \textbf{100} & 34.21 & 69.23 & 41.51 & 3.255\\
\addlinespace[2pt]
\shortstack[l]{GraF-VTA  + no memory} & 53.91 & \textbf{100} & 34.21 & 76.92 & 47.17 & \textbf{2.033}\\
\shortstack[l]{GraF-VTA + standard memory} & 61.72 & \textbf{100} & \textbf{42.11} & 76.92 & 60.38 & 2.247\\
\shortstack[l]{GraF-VTA  + pyramid memory} & \textbf{62.50} & \textbf{100} & \textbf{42.11} & \textbf{88.46} & 56.60 & 2.323\\
\shortstack[l]{GraF-VTA (dual graph) + pyramid memory} & \textbf{62.50} & \textbf{100} & 36.84 & 84.62 & \textbf{62.26} & 2.478\\
\bottomrule
\end{tabular*}
\end{table*}

\subsection{Ablation study}
\label{sec:panda_ablation}
We conduct the ablations in Table~\ref{tab:panda_memory_comparison} to assess how observation memory affects task success and inference time, and whether dual-graph conditioning provides additional benefits for partial-order tasks.

\vspace{-1.3em}
\paragraph{Observation memory improves task coverage.}
Compared with VTA, GraF-VTA achieves higher satisfaction rates on sequential and partial-order tasks, increasing from 69.23\% to 76.92\% and from 41.51\% to 47.17\%, respectively, while maintaining the same performance on multi-goal tasks. Observation memory further improves its performance on these more complex tasks. Specifically, standard memory improves satisfaction on multi-goal and partial-order tasks to 42.11\% and 60.38\%, respectively, while pyramid memory improves satisfaction across all three categories, reaching 42.11\%, 88.46\%, and 56.60\% on multi-goal, sequential, and partial-order tasks, respectively. Despite the additional inference time introduced by memory, both memory variants remain faster than VTA. Overall, these results demonstrate that observation memory further strengthens GraF-VTA's ability to generate satisfying trajectories for temporally structured tasks while maintaining efficient inference.

\vspace{-1.3em}
\paragraph{Dual-graph conditioning benefits partial-order tasks.}
With pyramid memory, dual-graph conditioning matches GraF-VTA's overall success rate and improves satisfaction on partial-order tasks. Satisfaction on multi-goal and sequential tasks decreases slightly relative to GraF-VTA with the same memory, but remains above the no-memory baseline. Inference time increases slightly, while remaining lower than VTA's. Additional results for the dual-graph variant are provided in the appendix.

\vspace{-1.3em}
\paragraph{Dual-view observations provide complementary information.}
The visual-input ablations in Figure~\ref{fig:figure2_panda_visual_stacked_single_column} suggest that complementary viewpoints remain important even with pyramid memory, as temporal context does not fully compensate for removing either view. The advantage of dual-view conditioning also persists when the single-view variants receive substantially larger sampling budgets, suggesting that access to complementary scene information helps beyond generating more candidates. These results establish that both views are necessary to sustain the task satisfaction performance of GraF-VTA with pyramid memory.

\subsection{Visual Grounding in Open-Loop Planning}
\label{sec:open_loop_comparison}
\begin{figure*}[t]
    \centering
    \includegraphics[width=\textwidth]{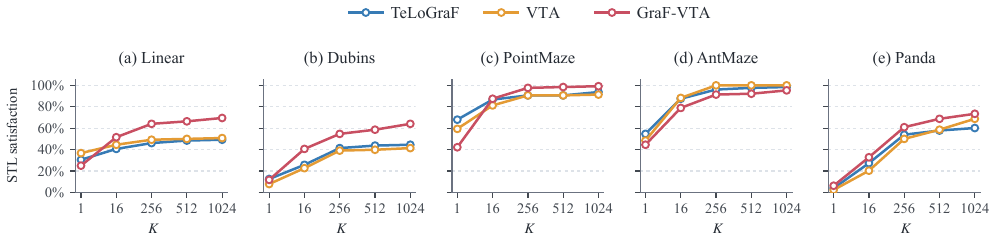}
    \caption{\textbf{Open-loop trajectory coverage.} GraF-VTA, using only the STL syntax graph, outperforms TeLoGraF and VTA in Success@$K$ on Linear, Dubins, PointMaze, and Panda at every plotted budget $K\geq16$.}
    \label{fig:openloop_visual_grounding}
\end{figure*}

\begin{figure*}[t]
    \centering
    \includegraphics[width=\textwidth]{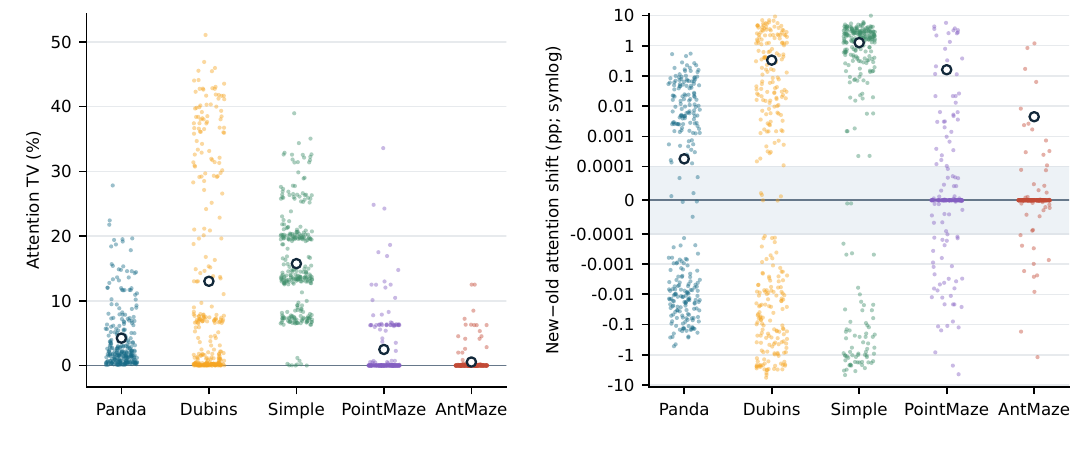}
    \caption{\textbf{Visual attention under object-ID swaps.} We swap IDs 2 and 3 in the STL graph while keeping other inputs fixed. Results cover 1,231 AP nodes from 549 eligible tasks across five environments, averaging attention over all fusion layers and heads. Left: attention total variation (TV). Right: change in new–old object attention preference (percentage points; symlog). Dots represent AP nodes; white circles indicate AP-equal arithmetic means. Linear (Simple) shows the strongest redistribution and target-directed shift, followed by Dubins. PointMaze exhibits a smaller positive mean shift, while Panda and AntMaze have near-zero mean shifts.}
    \label{fig:Visual_attention}
\end{figure*}

We compare open-loop trajectory generation by TeLoGraF, VTA, and GraF-VTA to evaluate the benefit of visual conditioning and the effectiveness of GraF-VTA's syntax-tree encoding. Figure~\ref{fig:openloop_visual_grounding} demonstrate that visual conditioning enables effective open-loop STL planning without requiring exact object geometry. For example, at $K=1024$, GraF-VTA improves over TeLoGraF by 20.31 and 19.53 percentage points on Linear and Dubins, respectively. At every plotted budget $K\geq16$, GraF-VTA outperforms both TeLoGraF and VTA in all environments except AntMaze, supporting the effectiveness of its syntax-tree encoding. GraF-VTA also exceeds both baselines' coverage at $K=1024$ using only $1/64$ of that sampling budget on Linear and $1/4$ on Dubins and PointMaze. On AntMaze, however, GraF-VTA performs slightly worse than VTA, and incomplete fixes to this environment cannot be ruled out as a contributing factor.

Figure~\ref{fig:Visual_attention} provides evidence that GraF-VTA uses symbolic object references to modulate visual attention. With visual observations held fixed, swapping object identities in the STL graph changes the attention distribution and can shift attention toward newly referenced objects. This behavior supports learned associations between atomic predicates and visual entities through logic–vision fusion, allowing visual attention to respond to the task specification. These results support task-dependent visual grounding in GraF-VTA, while showing that the strength and consistency of object alignment vary across environments.

\setlength{\openlooptablewidth}{0.8\textwidth}

\subsection{Execution coverage analysis across STL Types}
\label{sec:vta_dynamics_comparison}
\begin{figure*}[!t]
    \centering
    \includegraphics[width=\textwidth]{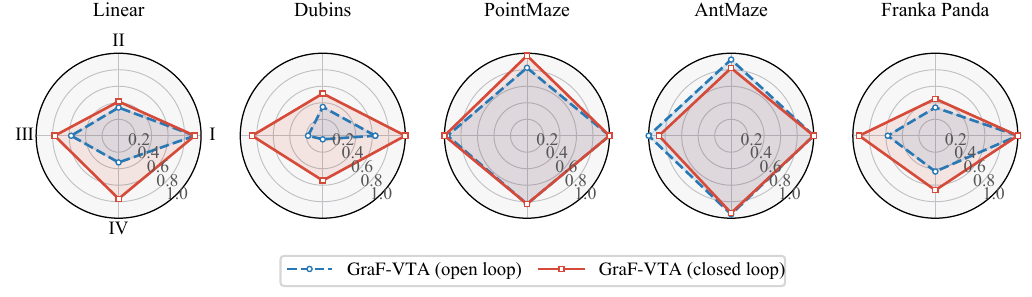}
    \caption{STL satisfaction rates of open-loop and closed-loop GraF-VTA with dynamics execution across five environments and four task categories at $K=1024$ (I: single-goal; II: multi-goal; III: sequential; IV: partial-order).}
    \label{fig:p2_radar}
\end{figure*}

\begin{table*}[!t]
\centering
\caption{\textbf{Closed-loop planning on Dubins.} GraF-VTA, using only the STL syntax graph, outperforms S-MSP and Logic-VLA in Success@$K$ and category-wise STL satisfaction at every reported candidate budget. Bold indicates the best result within each $K$ group.}
\label{tab:closed_loop_method_comparison}
\fontencoding{T1}\selectfont
\small

\setlength{\tabcolsep}{4pt}
\renewcommand{\arraystretch}{1.1}
\setlength{\extrarowheight}{0pt}
\setlength{\aboverulesep}{1.5pt}
\setlength{\belowrulesep}{1.5pt}
\begin{tabular*}{\openlooptablewidth}
{@{\extracolsep{\fill}}clc*{4}{>{\centering\arraybackslash}p{32pt}}@{}}
\toprule
\multirow{2}{*}{$K$}
& \multirow{2}{*}{Method}
& \multirow{2}{*}{\shortstack{Success@$K$\\(\%) $\uparrow$}}
& \multicolumn{4}{c}{STL satisfaction rate (\%) $\uparrow$} \\
\cmidrule(lr){4-7}
& & & I & II & III & IV \\
\midrule

\multirow{3}{*}{1}
& S-MSP
& 0.78 & 0.00 & 2.33 & 0.00 & 0.00  \\
& Logic-VLA
& 1.56 & 0.00 & 4.65 & 0.00 & 0.00  \\
& GraF-VTA 
& \textbf{17.19} & \textbf{45.45} & \textbf{11.63} & \textbf{28.57} & \textbf{8.70} \\
\midrule

\multirow{3}{*}{128}
& S-MSP
& 7.81 & 9.09 & 13.95 & 7.14 & 2.17 \\
& Logic-VLA
& 14.06 & 45.45 & 25.58 & 3.57 & 2.17  \\
& GraF-VTA 
& \textbf{51.56} & \textbf{90.91} & \textbf{37.21} & \textbf{82.14} & \textbf{36.96}  \\
\midrule

\multirow{3}{*}{256}
& S-MSP
& 9.38 & 18.18 & 16.28 & 7.14 & 2.17  \\
& Logic-VLA
& 17.97 & 45.45 & 30.23 & 10.71 & 4.35 \\
& GraF-VTA 
& \textbf{56.25} & \textbf{90.91} & \textbf{44.19} & \textbf{85.71} & \textbf{41.30}  \\
\midrule

\multirow{3}{*}{512}
& S-MSP
& 10.94 & 18.18 & 20.93 & 7.14 & 2.17  \\
& Logic-VLA
& 21.09 & 54.55 & 32.56 & 14.29 & 6.52  \\
& GraF-VTA 
& \textbf{62.50} & \textbf{100.00} & \textbf{46.51}
& \textbf{85.71} & \textbf{54.35}  \\
\midrule

\multirow{3}{*}{1024}
& S-MSP
& 12.50 & 18.18 & 23.26 & 10.71 & 2.17 \\
& Logic-VLA
& 27.34 & 63.64 & 44.19 & 17.86 & 8.70  \\
& GraF-VTA 
& \textbf{64.06}
& \textbf{100.00}
& \textbf{51.16}
& \textbf{85.71}
& \textbf{54.35}
 \\
\bottomrule
\end{tabular*}
\end{table*}

We compare open-loop and closed-loop dynamics execution of GraF-VTA across four STL task types at $K=1024$. As shown in Figure~\ref{fig:p2_radar}, closed-loop execution achieves higher overall coverage than open-loop execution in all environments except AntMaze. Specifically, Success@1024 increases from 46.88\% to 67.19\% in Linear, from 22.66\% to 64.06\% in Dubins, from 86.72\% to 91.41\% in PointMaze, and from 48.44\% to 67.97\% in Panda. The advantage on complex tasks is consistent across Linear, Dubins, and Panda. Closed-loop execution improves coverage on multi-goal, sequential, and partial-order tasks in all three environments, with larger gains for sequential and partial-order tasks than for multi-goal tasks. In AntMaze, closed-loop coverage remains high across all four task types despite a modest overall decrease relative to open-loop execution; incomplete environment fixes remain a possible but unverified explanation. These results support the effectiveness of closed-loop execution with updated visual observations and robot states in improving task success rates.

\subsection{Comparison with Other Closed-Loop Methods}
\label{sec:dubins_comparison}

\begin{figure}[!t]
    \centering
    \includegraphics[width=\linewidth]{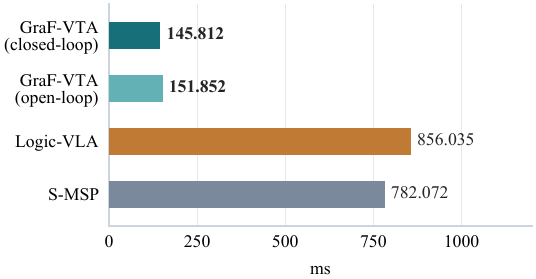}
    \caption{\textbf{Planning efficiency on Dubins.} GraF-VTA, configured with only the STL syntax graph, achieves lower median single-candidate first-decision planning latency than S-MSP and Logic-VLA under a separate controlled timing setup. All methods use FP32, with 10 flow steps for both flow models.}
    \label{fig:dubins_planning_latency}
\end{figure}

We compare GraF-VTA with S-MSP and Logic-VLA, which likewise use image observations and temporal-logic specifications, in closed-loop execution on Dubins. Table~\ref{tab:closed_loop_method_comparison} shows that GraF-VTA achieves higher overall success and STL satisfaction rates across all task categories. At $K=1024$, GraF-VTA improves Success@$K$ by 51.56 and 36.72 percentage points over S-MSP and Logic-VLA, respectively, while its category-wise satisfaction gains over Logic-VLA range from 6.97 to 67.85 percentage points. Figure~\ref{fig:dubins_planning_latency} shows $5.36\times$ and $5.87\times$ speedups in median single-candidate first-decision planning over S-MSP and Logic-VLA, respectively. These results demonstrate that GraF-VTA outperforms both comparable methods in task satisfaction and planning efficiency.

\FloatBarrier
\section{Conclusion and Limitations}
We presented GraF-VTA, which extends Vision-TL-Action with structured graph encoding, causal visual history, and action-prefix replanning for STL-conditioned trajectory generation. A temporal pyramid retains current visual detail and compresses past observations, while an optional frozen automaton embedding provides complementary untimed task structure. The single-graph configuration improves open-loop success coverage over the compared baselines in four of five simulated environments. Matched-update Panda ablations support the value of observation history, and Dubins comparisons demonstrate higher closed-loop coverage than adapted S-MSP and Logic-VLA.

Our evaluation uses simulated environments and semantic visual inputs, with gains varying across environments and no improvement on AntMaze. Success@$K$ measures candidate coverage rather than reliable online selection or guaranteed STL satisfaction. Finite observation history and a static automaton embedding do not provide complete task-progress monitoring. The dual-graph extension remains insufficiently evaluated, and the pyramid's additional gain over standard memory is modest. Future work will prioritize controlled component studies, replication across training seeds, and validation under real-world sensing and execution.

{
    \small
    \bibliographystyle{ieeenat_fullname}
    \bibliography{references}

@ARTICLE{athalye2026p2p,
  author={Athalye, Ashay and Kumar, Nishanth and Silver, Tom and Liang, Yichao and Wang, Jiuguang and Lozano-Pérez, Tomás and Kaelbling, Leslie Pack},
  journal={IEEE Robotics and Automation Letters}, 
  title={From Pixels to Predicates: Learning Symbolic World Models via Pretrained VLMs}, 
  year={2026},
  volume={11},
  number={4},
  pages={4002-4009},
}

@InProceedings{meng2025telograf,
  title = 	 {{T}e{L}o{G}ra{F}: Temporal Logic Planning via Graph-encoded Flow Matching},
  author =       {Meng, Yue and Fan, Chuchu},
  booktitle = 	 {Proceedings of the 42nd International Conference on Machine Learning},
  pages = 	 {43754--43780},
  year = 	 {2025},
  volume = 	 {267},
  series = 	 {Proceedings of Machine Learning Research},
}

@inproceedings{maler2004monitoring,
  title={Monitoring temporal properties of continuous signals},
  author={Maler, Oded and Nickovic, Dejan},
  booktitle={International symposium on formal techniques in real-time and fault-tolerant systems},
  pages={152--166},
  year={2004},
}

@InProceedings{zitkovich2023rt2,
  title = 	 {RT-2: Vision-Language-Action Models Transfer Web Knowledge to Robotic Control},
  author =       {Zitkovich, Brianna and Yu, Tianhe and Xu, Sichun and Xu, Peng and Xiao, Ted and Xia, Fei and Wu, Jialin and Wohlhart, Paul and Welker, Stefan and Wahid, Ayzaan and Vuong, Quan and Vanhoucke, Vincent and Tran, Huong and Soricut, Radu and Singh, Anikait and Singh, Jaspiar and Sermanet, Pierre and Sanketi, Pannag R. and Salazar, Grecia and Ryoo, Michael S. and Reymann, Krista and Rao, Kanishka and Pertsch, Karl and Mordatch, Igor and Michalewski, Henryk and Lu, Yao and Levine, Sergey and Lee, Lisa and Lee, Tsang-Wei Edward and Leal, Isabel and Kuang, Yuheng and Kalashnikov, Dmitry and Julian, Ryan and Joshi, Nikhil J. and Irpan, Alex and Ichter, Brian and Hsu, Jasmine and Herzog, Alexander and Hausman, Karol and Gopalakrishnan, Keerthana and Fu, Chuyuan and Florence, Pete and Finn, Chelsea and Dubey, Kumar Avinava and Driess, Danny and Ding, Tianli and Choromanski, Krzysztof Marcin and Chen, Xi and Chebotar, Yevgen and Carbajal, Justice and Brown, Noah and Brohan, Anthony and Arenas, Montserrat Gonzalez and Han, Kehang},
  booktitle = 	 {Proceedings of The 7th Conference on Robot Learning},
  pages = 	 {2165--2183},
  year = 	 {2023},
  volume = 	 {229},
  series = 	 {Proceedings of Machine Learning Research},
}

@InProceedings{kim2025openvla,
  title = 	 {OpenVLA: An Open-Source Vision-Language-Action Model},
  author =       {Kim, Moo Jin and Pertsch, Karl and Karamcheti, Siddharth and Xiao, Ted and Balakrishna, Ashwin and Nair, Suraj and Rafailov, Rafael and Foster, Ethan P and Sanketi, Pannag R and Vuong, Quan and Kollar, Thomas and Burchfiel, Benjamin and Tedrake, Russ and Sadigh, Dorsa and Levine, Sergey and Liang, Percy and Finn, Chelsea},
  booktitle = 	 {Proceedings of The 8th Conference on Robot Learning},
  pages = 	 {2679--2713},
  year = 	 {2025},
  editor = 	 {Agrawal, Pulkit and Kroemer, Oliver and Burgard, Wolfram},
  volume = 	 {270},
  series = 	 {Proceedings of Machine Learning Research},
}

@inproceedings{black2024pi0,
  title = {$\pi_0$: A Vision-Language-Action Flow Model for General Robot Control},
  author={Black, Kevin and Brown, Noah and Driess, Danny and Esmail, Adnan and Equi, Michael and Finn, Chelsea and Fusai, Niccolo and Groom, Lachy and Hausman, Karol and Ichter, Brian and others},
  booktitle = {RSS},
  year = {2025}
}

@inproceedings{raman2015reactive,
  title={Reactive synthesis from signal temporal logic specifications},
  author={Raman, Vasumathi and Donz{\'e}, Alexandre and Sadigh, Dorsa and Murray, Richard M and Seshia, Sanjit A},
  booktitle={Proceedings of the 18th international conference on hybrid systems: Computation and control},
  pages={239--248},
  year={2015}
}

@article{leung2023backpropagation,
  title={Backpropagation through signal temporal logic specifications: Infusing logical structure into gradient-based methods},
  author={Leung, Karen and Ar{\'e}chiga, Nikos and Pavone, Marco},
  journal={The International Journal of Robotics Research},
  volume={42},
  number={6},
  pages={356--370},
  year={2023},
}

@InProceedings{guo2024specification,
  title = 	 {Temporal Logic Specification-Conditioned Decision Transformer for Offline Safe Reinforcement Learning},
  author =       {Guo, Zijian and Zhou, Weichao and Li, Wenchao},
  booktitle = 	 {Proceedings of the 41st International Conference on Machine Learning},
  pages = 	 {17003--17019},
  year = 	 {2024},
  volume = 	 {235},
  series = 	 {Proceedings of Machine Learning Research},
}

@article{meng2024diverse,
  title={Diverse controllable diffusion policy with signal temporal logic},
  author={Meng, Yue and Fan, Chuchu},
  journal={IEEE Robotics and Automation Letters},
  volume={9},
  number={10},
  pages={8354--8361},
  year={2024},
}

@article{kapoor2025stlcg,
  title={Stlcg++: A masking approach for differentiable signal temporal logic specification},
  author={Kapoor, Parv and Mizuta, Kazuki and Kang, Eunsuk and Leung, Karen},
  journal={IEEE Robotics and Automation Letters},
  volume={10},
  number={9},
  pages={9240--9247},
  year={2025},
}

@article{feng2024ltldog,
  title   = {{LTLDoG}: Satisfying Temporally-Extended Symbolic Constraints for Safe Diffusion-Based Planning},
  author  = {Feng, Zeyu and Luan, Hao and Goyal, Pranav and Soh, Harold},
  journal = {IEEE Robotics and Automation Letters},
  volume  = {9},
  number  = {10},
  pages   = {8571--8578},
  year    = {2024},
}

@article{liu2026zeroshotstl,
  title={Zero-shot trajectory planning for signal temporal logic tasks},
  author={Liu, Ruijia and Hou, Ancheng and Yu, Xiao and Yin, Xiang},
  journal={Advances in Neural Information Processing Systems},
  volume={38},
  pages={130405--130442},
   year = {2025}
}

@article{chi2025diffusion,
  title={Diffusion policy: Visuomotor policy learning via action diffusion},
  author={Chi, Cheng and Xu, Zhenjia and Feng, Siyuan and Cousineau, Eric and Du, Yilun and Burchfiel, Benjamin and Tedrake, Russ and Song, Shuran},
  journal={The International Journal of Robotics Research},
  volume={44},
  number={10-11},
  pages={1684--1704},
  year={2025},
}

@inproceedings{zhang2025flowpolicy,
  title={Flowpolicy: Enabling fast and robust 3d flow-based policy via consistency flow matching for robot manipulation},
  author={Zhang, Qinglun and Liu, Zhen and Fan, Haoqiang and Liu, Guanghui and Zeng, Bing and Liu, Shuaicheng},
  booktitle={Proceedings of the AAAI Conference on Artificial Intelligence},
  volume={39},
  number={14},
  pages={14754--14762},
  year={2025}
}

@inproceedings{Kamath_2021_ICCV,
  title={Mdetr-modulated detection for end-to-end multi-modal understanding},
  author={Kamath, Aishwarya and Singh, Mannat and LeCun, Yann and Synnaeve, Gabriel and Misra, Ishan and Carion, Nicolas},
  booktitle={2021 IEEE/CVF international conference on computer vision (ICCV)},
  pages={1760--1770},
  year={2021},
}

@inproceedings{Li_2022_CVPR,
  title={Grounded language-image pre-training},
  author={Li, Liunian Harold and Zhang, Pengchuan and Zhang, Haotian and Yang, Jianwei and Li, Chunyuan and Zhong, Yiwu and Wang, Lijuan and Yuan, Lu and Zhang, Lei and Hwang, Jenq-Neng and others},
  booktitle={2022 IEEE/CVF Conference on Computer Vision and Pattern Recognition (CVPR)},
  pages={10955--10965},
  year={2022},
}

@INPROCEEDINGS{ye2026smsp,
  title={Bridging perception and planning: Towards end-to-end planning for signal temporal logic tasks},
  author={Ye, Bowen and Huang, Junyue and Liu, Yang and Qiao, Xiaozhen and Yin, Xiang},
  booktitle={2026 IEEE International Conference on Robotics and Automation (ICRA)},
  pages={5054--5061},
  year={2026},
}

@article{belta2019formal,
  title={Formal methods for control synthesis: An optimization perspective},
  author={Belta, Calin and Sadraddini, Sadra},
  journal={Annual Review of Control, Robotics, and Autonomous Systems},
  volume={2},
  number={1},
  pages={115--140},
  year={2019},
}

@inproceedings{vasile2017sampling,
  title={Sampling-based synthesis of maximally-satisfying controllers for temporal logic specifications},
  author={Vasile, Cristian-Ioan and Raman, Vasumathi and Karaman, Sertac},
  booktitle={2017 IEEE/RSJ International Conference on Intelligent Robots and Systems (IROS)},
  pages={3840--3847},
  year={2017},
}

@inproceedings{sadraddini2015robust,
  title={Robust temporal logic model predictive control},
  author={Sadraddini, Sadra and Belta, Calin},
  booktitle={2015 53rd Annual Allerton Conference on Communication, Control, and Computing (Allerton)},
  pages={772--779},
  year={2015},
}

@inproceedings{dawson2022robust,
  title={Robust counterexample-guided optimization for planning from differentiable temporal logic},
  author={Dawson, Charles and Fan, Chuchu},
  booktitle={2022 IEEE/RSJ International Conference on Intelligent Robots and Systems (IROS)},
  pages={7205--7212},
  year={2022},
  organization={IEEE}
}

@article{meng2023signal,
  title={Signal temporal logic neural predictive control},
  author={Meng, Yue and Fan, Chuchu},
  journal={IEEE Robotics and Automation Letters},
  volume={8},
  number={11},
  pages={7719--7726},
  year={2023},
}

@InProceedings{janner2022planning,
  title = 	 {Planning with Diffusion for Flexible Behavior Synthesis},
  author =       {Janner, Michael and Du, Yilun and Tenenbaum, Joshua and Levine, Sergey},
  booktitle = 	 {Proceedings of the 39th International Conference on Machine Learning},
  pages = 	 {9902--9915},
  year = 	 {2022},
  volume = 	 {162},
  series = 	 {Proceedings of Machine Learning Research},
}

@inproceedings{zhong2023guided,
  title={Guided conditional diffusion for controllable traffic simulation},
  author={Zhong, Ziyuan and Rempe, Davis and Xu, Danfei and Chen, Yuxiao and Veer, Sushant and Che, Tong and Ray, Baishakhi and Pavone, Marco},
  booktitle={2023 IEEE international conference on robotics and automation (ICRA)},
  pages={3560--3566},
  year={2023},
}

@article{brohan2023rt,
  title={Rt-1: Robotics transformer for real-world control at scale},
  author={Brohan, Anthony and Brown, Noah and Carbajal, Justice and Chebotar, Yevgen and Dabis, Joseph and Finn, Chelsea and Gopalakrishnan, Keerthana and Hausman, Karol and Herzog, Alex and Hsu, Jasmine and others},
  journal={arXiv preprint arXiv:2212.06817},
  year={2022}
}

@inproceedings{openx2024,
  title={Open x-embodiment: Robotic learning datasets and rt-x models: Open x-embodiment collaboration 0},
  author={O’Neill, Abby and Rehman, Abdul and Maddukuri, Abhiram and Gupta, Abhishek and Padalkar, Abhishek and Lee, Abraham and Pooley, Acorn and Gupta, Agrim and Mandlekar, Ajay and Jain, Ajinkya and others},
  booktitle={2024 IEEE International Conference on Robotics and Automation (ICRA)},
  pages={6892--6903},
  year={2024},
}

@inproceedings{li2024roboflamingo,
  title={Vision-language foundation models as effective robot imitators},
  author={Li, Xinghang and Liu, Minghuan and Zhang, Hanbo and Yu, Cunjun and Xu, Jie and Wu, Hongtao and Cheang, Chilam and Jing, Ya and Zhang, Weinan and Liu, Huaping and others},
  booktitle={International Conference on Learning Representations},
  volume={2024},
  pages={26703--26721},
  year={2024}
}

@article{ghosh2024octo,
  title={Octo: An open-source generalist robot policy},
  author={Team, Octo Model and Ghosh, Dibya and Walke, Homer and Pertsch, Karl and Black, Kevin and Mees, Oier and Dasari, Sudeep and Hejna, Joey and Kreiman, Tobias and Xu, Charles and others},
  journal={arXiv preprint arXiv:2405.12213},
  year={2024}
}

@article{bjorck2025groot,
  title={Gr00t n1: An open foundation model for generalist humanoid robots},
  author={Bjorck, Johan and Casta{\~n}eda, Fernando and Cherniadev, Nikita and Da, Xingye and Ding, Runyu and Fan, Linxi and Fang, Yu and Fox, Dieter and Hu, Fengyuan and Huang, Spencer and others},
  journal={arXiv preprint arXiv:2503.14734},
  year={2025}
}

@inproceedings{ye2026starvlaalpha,
  title={StarVLA-$\alpha$: Reducing complexity in vision-language-action systems},
  author={Ye, Jinhui and Gao, Ning and Yang, Senqiao and Zheng, Jinliang and Wang, Zixuan and Chen, Yuxin and Chen, Pengguang and Chen, Yilun and Liu, Shu and Jia, Jiaya},
  booktitle={European Conference on Computer Vision},
  pages={351--369},
  year={2026},
}

@inproceedings{pi2025pi05,
  title = 	 {$\pi_{0.5}$: a Vision-Language-Action Model with Open-World Generalization},
  author =       {Black, Kevin and Brown, Noah and Darpinian, James and Dhabalia, Karan and Driess, Danny and Esmail, Adnan and Equi, Michael Robert and Finn, Chelsea and Fusai, Niccolo and Galliker, Manuel Y. and Ghosh, Dibya and Groom, Lachy and Hausman, Karol and ichter, brian and Jakubczak, Szymon and Jones, Tim and Ke, Liyiming and LeBlanc, Devin and Levine, Sergey and Li-Bell, Adrian and Mothukuri, Mohith and Nair, Suraj and Pertsch, Karl and Ren, Allen Z. and Shi, Lucy Xiaoyang and Smith, Laura and Springenberg, Jost Tobias and Stachowicz, Kyle and Tanner, James and Vuong, Quan and Walke, Homer and Walling, Anna and Wang, Haohuan and Yu, Lili and Zhilinsky, Ury},
  booktitle = 	 {Proceedings of The 9th Conference on Robot Learning},
  pages = 	 {17--40},
  year = 	 {2025},
  volume = 	 {305},
  series = 	 {Proceedings of Machine Learning Research},
 
}

@inproceedings{zhen2024threedvla,
  title = 	 {3{D}-{VLA}: A 3{D} Vision-Language-Action Generative World Model},
  author =       {Zhen, Haoyu and Qiu, Xiaowen and Chen, Peihao and Yang, Jincheng and Yan, Xin and Du, Yilun and Hong, Yining and Gan, Chuang},
  booktitle = 	 {Proceedings of the 41st International Conference on Machine Learning},
  pages = 	 {61229--61245},
  year = 	 {2024},
  volume = 	 {235},
  series = 	 {Proceedings of Machine Learning Research},
}

@article{qu2025spatialvla,
  title={Spatialvla: Exploring spatial representations for visual-language-action model},
  author={Qu, Delin and Song, Haoming and Chen, Qizhi and Yao, Yuanqi and Ye, Xinyi and Ding, Yan and Wang, Zhigang and Gu, JiaYuan and Zhao, Bin and Wang, Dong and others},
  journal={arXiv preprint arXiv:2501.15830},
  year={2025}
}

@inproceedings{feng2025diffusion,
  title={Diffusion meets options: Hierarchical generative skill composition for temporally-extended tasks},
  author={Feng, Zeyu and Luan, Hao and Ma, Kevin Yuchen and Soh, Harold},
  booktitle={2025 IEEE International Conference on Robotics and Automation (ICRA)},
  pages={10854--10860},
  year={2025},
}

@inproceedings{vaezipoor2021ltl2action,
  title={Ltl2action: Generalizing ltl instructions for multi-task rl},
  author={Vaezipoor, Pashootan and Li, Andrew C and Icarte, Rodrigo A Toro and Mcilraith, Sheila A},
  booktitle={International Conference on Machine Learning},
  pages={10497--10508},
  year={2021},
}

@article{lindemann2021funnel,
  title={Funnel control for fully actuated systems under a fragment of signal temporal logic specifications},
  author={Lindemann, Lars and Dimarogonas, Dimos V},
  journal={Nonlinear Analysis: Hybrid Systems},
  volume={39},
  pages={100973},
  year={2021},
}

@article{lindemann2019feedback,
  title={Feedback control strategies for multi-agent systems under a fragment of signal temporal logic tasks},
  author={Lindemann, Lars and Dimarogonas, Dimos V},
  journal={Automatica},
  volume={106},
  pages={284--293},
  year={2019},
}

@InProceedings{jiang2023vima,
  title = 	 {{VIMA}: Robot Manipulation with Multimodal Prompts},
  author =       {Jiang, Yunfan and Gupta, Agrim and Zhang, Zichen and Wang, Guanzhi and Dou, Yongqiang and Chen, Yanjun and Fei-Fei, Li and Anandkumar, Anima and Zhu, Yuke and Fan, Linxi},
  booktitle = 	 {Proceedings of the 40th International Conference on Machine Learning},
  pages = 	 {14975--15022},
  year = 	 {2023},
  volume = 	 {202},
  series = 	 {Proceedings of Machine Learning Research},
}

@inproceedings{liu2024grounding,
  title={Grounding dino: Marrying dino with grounded pre-training for open-set object detection},
  author={Liu, Shilong and Zeng, Zhaoyang and Ren, Tianhe and Li, Feng and Zhang, Hao and Yang, Jie and Jiang, Qing and Li, Chunyuan and Yang, Jianwei and Su, Hang and others},
  booktitle={European conference on computer vision},
  pages={38--55},
  year={2024},
}

@inproceedings{jackermeier2025deepltl,
  title={Deepltl: Learning to efficiently satisfy complex ltl specifications for multi-task rl},
  author={Jackermeier, Mathias and Abate, Alessandro},
  booktitle={International Conference on Learning Representations},
  volume={2025},
  pages={14000--14028},
  year={2025}
}

@article{liu2026visiontlaction,
  title={Vision-tl-action: Neuro-symbolic trajectory generation from visual observations and temporal logic},
  author={Liu, Zezhi and Zheng, Zhiwei and Luo, Hanqian and Qin, Deyun and Wu, Shizhen and Fang, Yongchun},
  journal={arXiv preprint arXiv:2607.26770},
  year={2026}
}

@article{yalcinkaya2024compositional,
  title={Compositional automata embeddings for goal-conditioned reinforcement learning},
  author={Yalcinkaya, Beyazit and Lauffer, Niklas and Vazquez-Chanlatte, Marcell and Seshia, Sanjit},
  journal={Advances in Neural Information Processing Systems},
  volume={37},
  pages={72933--72963},
  year={2024}
}

@article{fu2020d4rl,
  title = {{D4RL}: Datasets for Deep Data-Driven Reinforcement Learning},
  author = {Fu, Justin and Kumar, Aviral and Nachum, Ofir and Tucker, George and Levine, Sergey},
  journal = {arXiv preprint arXiv:2004.07219},
  year = {2020},
  url = {https://arxiv.org/abs/2004.07219}
}

@article{gaz2019dynamic,
  title={Dynamic identification of the franka emika panda robot with retrieval of feasible parameters using penalty-based optimization},
  author={Gaz, Claudio and Cognetti, Marco and Oliva, Alexander and Giordano, Paolo Robuffo and De Luca, Alessandro},
  journal={IEEE Robotics and Automation Letters},
  volume={4},
  number={4},
  pages={4147--4154},
  year={2019},
}

@inproceedings{lipman2023flow,
  title = {Flow Matching for Generative Modeling},
  author = {Lipman, Yaron and Chen, Ricky T. Q. and Ben-Hamu, Heli and Nickel, Maximilian and Le, Matt},
  booktitle = {International Conference on Learning Representations (ICLR)},
  year = {2023},
}

@article{wang2026logicvla,
  title={Logic-vla: A temporal logic conditioned vision-language-action model},
  author={Wang, Celina Shiyu and Zhao, Yiqi and Ye, Junjie and Wang, Yue and Deshmukh, Jyotirmoy V},
  journal={arXiv preprint arXiv:2608.20556},
  year={2026}
}

@inproceedings{he2016deep,
  title={Deep residual learning for image recognition},
  author={He, Kaiming and Zhang, Xiangyu and Ren, Shaoqing and Sun, Jian},
  booktitle={Proceedings of the IEEE conference on computer vision and pattern recognition},
  pages={770--778},
  year={2016}
}

@article{levine2016endtoend,
  title={End-to-end training of deep visuomotor policies},
  author={Levine, Sergey and Finn, Chelsea and Darrell, Trevor and Abbeel, Pieter},
  journal={Journal of Machine Learning Research},
  volume={17},
  number={39},
  pages={1--40},
  year={2016}
}

@inproceedings{xie2021embedding,
  title={Embedding symbolic temporal knowledge into deep sequential models},
  author={Xie, Yaqi and Zhou, Fan and Soh, Harold},
  booktitle={2021 IEEE International Conference on Robotics and Automation (ICRA)},
  pages={4267--4273},
  year={2021},
}

@inproceedings{shi2026memoryvla,
  title={Memoryvla: Perceptual-cognitive memory in vision-language-action models for robotic manipulation},
  author={Shi, Hao and Xie, Bin and Liu, Yingfei and Sun, Lin and Liu, Fengrong and Wang, Tiancai and Zhou, Erjin and Fan, Haoqiang and Zhang, Xiangyu and Huang, Gao},
  booktitle={International Conference on Learning Representations},
  volume={2026},
  pages={18567--18602},
  year={2026}
}

@inproceedings{koo2026hamlet,
author = {Koo, Myungkyu and Choi, Daewon and Kim, Taeyoung and Lee, Kyungmin and Kim, Changyeon and Seo, Younggyo and Shin, Jinwoo },
 booktitle = {International Conference on Learning Representations (ICLR)},
 pages = {101537--101558},
 title = {HAMLET: Switch Your Vision-Language-Action Model into a History-Aware Policy},
 year = {2026}
}

@article{torne2026mem,
  title={MEM: Multi-Scale Embodied Memory for Vision Language Action Models}, 
  author={Marcel Torne and Karl Pertsch and Homer Walke and Kyle Vedder and Suraj Nair and Brian Ichter and Allen Z. Ren and Haohuan Wang and Jiaming Tang and Kyle Stachowicz and Karan Dhabalia and Michael Equi and Quan Vuong and Jost Tobias Springenberg and Sergey Levine and Chelsea Finn and Danny Driess},
  journal = {arXiv preprint arXiv:2603.03596},
  year = {2026},
}
}

\end{document}